\documentclass[letterpaper, 10 pt, conference]{ieeeconf}  

\IEEEoverridecommandlockouts                              

\usepackage{amsmath} 
\usepackage{amssymb}  
\usepackage{graphicx}
\usepackage{bbm}
\usepackage{amsmath}
\usepackage{algorithm}
\usepackage{multirow} 
\usepackage{makecell}
\usepackage[flushleft]{threeparttable}
\usepackage{color}

\usepackage{caption}
\usepackage{subcaption}
\usepackage{hyperref}
\usepackage{algorithm}
\usepackage{graphicx} 
\usepackage{booktabs}  
\usepackage{tabularx}  
\usepackage{threeparttable}  
\usepackage{array}     
\usepackage{multirow}
\usepackage{algpseudocode}

\usepackage{amsthm}

\usepackage{cite}
\theoremstyle{plain}

\newtheorem{defn}{Definition}[section]

\usepackage{graphicx}
\newcommand{\indep}{\rotatebox[origin=c]{90}{$\models$}}

\title{\LARGE \bf
Causal Adjacency Learning for Spatiotemporal Prediction Over Graphs
}

\author{Zhaobin Mo$^{1,*}$, Qingyuan Liu$^{2,*}$, Baohua Yan$^{3}$, Longxiang Zhang$^{4}$ and Xuan Di$^{1,4, \dag}$
\thanks{This work is sponsored NSF SCC-2218809.}
\thanks{$^{*}$Equal contributions; $^{\dag}$Corresponding author;}
\thanks{$^{1}$These authors are affiliated with the Department of Civil Engineering and Engineering Mechanics, Columbia University, 500 West 120th Street, New York, NY 10025, USA
        {\tt\small zm2302@columbia.edu, sharon.di@columbia.edu }}%
\thanks{$^{2}$This author is affiliated with the Department of Electrical Engineering, Columbia University, 500 West 120th Street, New York, NY 10025, USA
        {\tt\small ql2505@columbia.edu  }}%
\thanks{$^{3}$This author is affiliated with the Department of Applied Physics and Applied Mathematics, Columbia University, 500 West 120th Street, New York, NY 10025, USA
        {\tt\small by2348@columbia.edu  }}%
\thanks{$^{4}$These authors are affiliated with the Data Science Institute, Columbia University, 550 W 120th St, New York, NY 10027, USA
        {\tt\small lz2869@columbia.edu, sharon.di@columbia.edu }}%
}

\begin{document}

\maketitle
\thispagestyle{empty}
\pagestyle{empty}

\begin{abstract}

Spatiotemporal prediction over graphs (STPG) is crucial for transportation systems. In existing STPG models, an adjacency matrix is an important component that captures the relations among nodes over graphs. However, most studies calculate the adjacency matrix by directly memorizing the data, such as distance- and correlation-based matrices. These adjacency matrices do not consider potential pattern shift for the test data, and may result in suboptimal performance if the test data has a different distribution from the training one. This issue is known as the Out-of-Distribution generalization problem. To address this issue, in this paper we propose a Causal Adjacency Learning (CAL) method to discover causal relations over graphs. The learned causal adjacency matrix is evaluated on a downstream spatiotemporal prediction task using real-world graph data. Results demonstrate that our proposed adjacency matrix can capture the causal relations, and using our learned adjacency matrix can enhance prediction performance on the OOD test data, even though causal learning is not conducted in the downstream task.
\end{abstract}

\section{Introduction}

Spatiotemporal (ST) prediction over graphs (STPG) aims to uncover the dynamics of graph-structured data along its temporal evolution. This method has been widely studied in several transportation-related areas, including predictions of traffic flow and vehicle speeds \cite{wang2023stgin,wang2023st}. Understanding these spatiotemporal patterns is crucial for making better decision, improving the allocation of resources, and strengthening risk management efforts \cite{zhuang2022robust,zhuang2022does,zhuang2022defending}.

In the STPG problem, an important question is how to calculate an adjacency matrix that captures the inter-nodal relations. Existing studies mainly employ three methods for calculating adjacency matrices: heuristic, correlation, and attention mechanisms. Heuristic methods \cite{yu2017spatio} use predefined human knowledge, such as the geospatial distance among nodes, assuming that neighboring nodes have a higher impact if they are closer to the current node. Similarly, correlation \cite{chen2023invariant} and attention-based \cite{bai2020adaptive,guo2019attention,wang2022event,lu2020spatiotemporal} methods assume that a neighboring node with a high impact will have a high correlation or attention score, respectively. Additionally, some studies propose a multi-graph method that combines different graphs. However, these methods often fail to consider potential shifts in the data, such as those caused by events like the COVID-19 outbreak. This issue is known as the out-of-distribution (OOD) generalization problem. Consequently, applying the learned adjacency matrix to OOD data may result in suboptimal performance.

To address the OOD issue, some studies have incorporated causal inference into STPG \cite{zhang2022dynamic,xia2024deciphering,ruan2024infostgcan}. Causal inference aims to identify causality, as opposed to merely correlation, among different variables. The underlying idea is that causal relations remain constant for OOD data, while correlation relations may not. However, most existing causal inference approaches for STPG underemphasize the temporal dimension. Additionally, most of these studies conduct causal inference in the latent space, where the causal effects of the nodes are represented as latent variables. This representation, however, lacks interpretability and is not transferable after learning.

To fill in this research gap, we propose a Causal Adjacency Learning (CAL) framework in this study to learn causal relationships and encode them into the adjacency matrix. Unlike existing causal inference methods, our proposed method considers the temporal dimension to calculate causal relations in the spatiotemporal prediction setting and provides a transferable upstream task of learning a causal adjacency matrix. Compared to other matrices, our learned adjacency matrix successfully encodes causal relations and can be transferred to a downstream task without the need to conduct causal inference again. This method is inspired by the temporal causal feature selection, which concerns about deciding if one time-series variable can decide another one. We extend this idea from the temporal domain to the spatiotemporal domain, and first apply it to propose a transferrable CAL framework for addressing the OOD issue.
Our contributions are:
\begin{itemize}
    \item We are the first to consider temporal dimensions in the problem of spatiotemporal prediction over graphs;
    \item We provide the transferrable upstream causal adjacency learning framework;
    \item Our proposed method is evaluated using real-world graph data with OOD patterns.
\end{itemize}

The rest of the paper is organized as follows: Section~\ref{sec:lit_rev} review the related work. Section~\ref{sec:preliminary_problem} introduces the preliminaries and problem statement. Section~\ref{sec:methodology} introduces the methodology of our proposed framework. Section~\ref{sec:exp} introduces the experiment and results using real-wi experiments and results. Section~\ref{sec:conclu} concludes and projects future research directions.

\section{Related Work}
\label{sec:lit_rev}
In this section, we introduce two important relevant topics in STPG, the adjacency matrix and causal inference in the setting of STPG.

\subsection{Adjacency Matrix}
An adjacency matrix is a crucial tool for leveraging the non-Euclidean nature of graph data. There are various types of adjacency matrices designed for different analytical purposes. Heuristic methods calculate adjacency matrices using human prior knowledge. For example, distance-based matrices emphasize geospatial relations among nodes. \cite{yu2017spatio} first calculate the inter-node distance by geographical coordinates, and set a threshold to determine the connectivity between each two nodes. Apart from the geographic distance, \cite{li2021spatial} uses road-network distance, which considers the actual path distance between each node. Additionally, some studies utilize road connections and contextual similarity in their matrices. Although these matrices have strong semantic significance, however, their main limitation is that they do not consider temporal dynamics. In contrast, correlation \cite{zhang2019trafficgan} and attention-based \cite{bai2020adaptive,guo2019attention,mo2024cross,mo2024pi} methods are more adaptive to the evolving feature and account for the temporal dimension by incorporating time series data to calculate the matrix. Nonetheless, these adjacency matrices mostly just memorize the pattern of the training data. Both the correlation and attention-based methods consider the temporal dimensions in a way that the temporal patterns are aggregated in the form of a correlation matrix or attention score. This aggregation will hide the potential pattern shift, which is the OOD issue.

\subsection{Causality-based Methods in STPG}
Causal methods can effectively address the OOD issue \cite{ruan2023causal}. The fundamental assumption behind causality-based methods is that the effect of some variables on others will remain constant for out-of-distribution data. Thus, these variables that hold constant relations are also known as causal factors. Causal-based methods are not new for graph applications. For example, \cite{chen2022learning} aims to identify the key protein subgraph as the causal factor to determine the functionality of the protein, while \cite{fan2022debiasing} identifies key pixels in images. However, most relevant studies do not consider the temporal dimension of a graph, and there are two relevant studies on causality-based methods for STPG. \cite{zhang2022dynamic} applied an attention mechanism to learn a feature mask that can identify the causal feature. An invariance loss function is designed to guide the model to generate the correct causal feature. While it considers the temporal dimension, the time-series data is used as a chunk to predict the causal chunk, and the patterns within the temporal dimension are understudied. In \cite{xia2024deciphering}, the causal factor is learned as a latent variable that inputs the spatiotemporal data chunk. This approach also treats the temporal dimension as a chunk without studying the patterns within it. Furthermore, learning a latent code as a causal factor lacks interpretability. Our paper differs from these studies in that we consider the within-temporal features, and we learn the causal factor as an adjacency matrix that can be easily transferred to downstream tasks.

\section{Priliminaries and Problem Statement}
\label{sec:preliminary_problem}
\subsection{Graph and Adjacency Matrix}
In an undirected graph \(\mathcal{G}=(V,E,A)\), \(V=\{v_i\}_{i=1}^{N}\) represents the nodes, with \(N=|V|\) being the total number of nodes. The edges are represented by the set \(E\), and the adjacency matrix \(A \in \mathbb{R}^{N \times N}\) denotes node connectivity. An entry \(A_{ij}\) in the matrix is set to 1 if an edge \((v_i, v_j)\) exists in \(E\), and 0 otherwise. Node features are encapsulated in \(X\). For example, the distance-based adjacency matrix is calculated as
\begin{equation}
    (A_{DIS})_{i j}= \begin{cases}\exp \left(-\frac{d_{i j}^2}{\sigma^2}\right), & i \neq j \text { and } \exp \left(-\frac{d_{i j}^2}{\sigma^2}\right) \geq \epsilon \\ 0, & \text { otherwise }
    \end{cases},
\end{equation}
where $d_{i,j}$ is the distance between nodes $i$ and $j$, $\sigma$ is a normalization parameter, and $\epsilon$ is a threshold that determines the connectivity among nodes based on their distance. Also, the correlation-based adjacency matrix is calculated as:
\begin{equation}
\label{eq:corr}
    (A_{COR})_{ij} = \frac{\sum_{t=1}^T(\mathbf{x}^{i}_t-\bar{\mathbf{x}}^{i}) (\mathbf{x}^{j}_t-\bar{\mathbf{x}}^{j} )}{\sqrt{\sum_{t=1}^T(\mathbf{x}^{i}_t-\bar{\mathbf{x}}^{i})^2} \sqrt{\sum_{t=1}^T(\mathbf{x}^{j}_t-\bar{\mathbf{x}}^{j})^2} },
\end{equation},
where $\mathbf{x}^{i}_t$ is the feature of node $i$ at time step $t$ and $\bar{\mathbf{x}}^{i}$ is its averaged value across time.

\subsection{Problem Statement}
With all the preliminaries introduced above, we are ready to define the problem of predicting human mobility, i.e., the future regional number of visits. Given the historical nodal feature of previous $\tau$ time window, $\mathbf{X}_{(t-\tau+1):t} = [ \mathbf{X}_{t-\tau+1},\cdots,\mathbf{X}_t ]$, we aim to learn a function $f$ to predict the future $\tau^{\prime}$-length mobility sequence $Y_{(t+1):(t+\tau^{\prime})}=[Y_{t+1}, \cdots,Y_{(t+1):(t+\tau^{\prime})}]$.

\begin{equation}
\label{eq:problem_state_1}
[ \mathbf{X}_{(t-\tau+1):t}; A ] \stackrel{f}{\rightarrow}Y_{(t+1):(t+\tau^{\prime})},
\end{equation}
where $\mathbf{X}_t =(\mathbf{x}^{1}_t,...,\mathbf{x}^{N}_t)$ represents all the nodal feature at time t; $A$ is the adjacency matrix; ${Y_t} = (y_t^{1}, \cdots y_t^{N})$ stands for all the prediction targets, with $y_t^{i} \in \mathbb{R}$ being the prediction target of node $i$ at time $t$, where $i \in \{1, \cdots, N\}$. Specifically, our goal will focus on the prediction performance of test data that has an apparent OOD pattern.

\section{Methodology}
\label{sec:methodology}
We will introduce the framework of our proposed framework for STPG which is illustrated in Fig. 1. The proposed method consists of two modules, an upstream CAL and a downstream spatiotemporal graph neural network (GNN). We make such a design so that the adjacency matrix $A_{CAU}$ that is learned from the upstream can be interpretable and transferrable for the downstream tasks. 
\begin{figure*}[htbp]
\centering
    \includegraphics[width=0.8\linewidth]{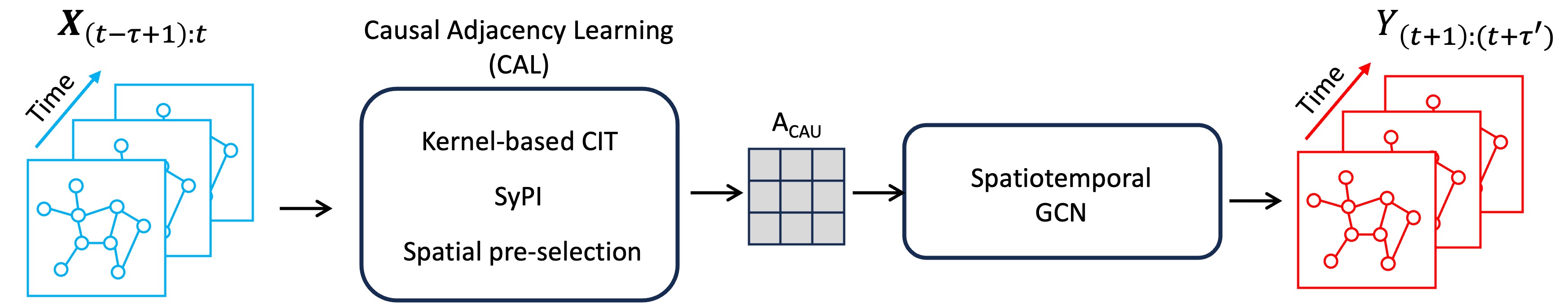}
    \caption{Framework of the upstream CAL and the downstream spatiotemporal GCN for the problem of STPG.}
    \label{fig:pipeline}
\end{figure*}

\subsection{Causal Adjacency Learning}
The core of our proposed framework is to learn the causal adjacency matrix that captures the causal relations among nodes. In the framework of CAL, we use the temporal conditional independence test (CIT) \cite{sen2017model,agterberg2002conditional,doran2014permutation} to identify the causal relations. We further incorporate spatial correlations into the temporal CIT framework to decrease the computational burden and also to encode the spatial relations. 

\subsubsection{Conditional Independence Test (CIT)}
An important tool for deciding the causal factor is the conditional independence test which has been applied to causal discovery. We define the concept of condition independence in the context of STPG as follows:
\begin{defn}{(Conditional Independence) }
    Let $\mathbf{x}^{i}_{1:t}$,  $\mathbf{x}^{j}_{1:t}$ and $\mathbf{x}^{k}_{1:t}$ denote time-series features of three nodes. The conditional independence is denoted as $\mathbf{x}^{i}_{1:t} \indep \mathbf{x}^{j}_{1:t} \mid \mathbf{x}^{k}_{1:t}$. This conditional independence indicates that given the time-series feature of node $k$, knowing the time-series feature of node $i$ (or $j$) does not provide additional information about node $j$ (or $i$).
\end{defn}
Conditional independence provides guidance for selecting causal features, given the fact that knowing the information of the causal feature can get rid of other features that may have a variant relation across data distribution (also known as environment features). As our problem is on continuous space, we adopt a kernel-based CIT to identify the causal features. The kernel-based method is used to approximate the distribution of a continuous variable.  The algorithm of kernel-based CIT is illustrated in Algorithm 1. Note that we omit some technical detail like normalization for notation simplicity for STPG. Readers can refer to \cite{zhang2012kernel} for the complete derivation of the kernel-based CIT.

\begin{algorithm}
\caption{Kernel-based CIT}
\begin{algorithmic}[1]
\State \textbf{Input:} $i$: A permutation of nodes time-series feature combination $(\mathbf{x}^{i}_{1:t},\mathbf{x}^{j}_{1:t},\mathbf{x}^{k}_{1:t})$ where $i \neq j \neq k \in \{1,...,N\}$; the significance level $\alpha$; the Gaussian Kernel $\mathbf{K}=\exp \left(-\frac{\left\|\mathbf{x}_1-\mathbf{x}_2\right\|^2}{2 \sigma_X^2}\right)$ with $\sigma_X$ being the kernel width.

\State Calculate centralized kernel matrice for the nodes feature permutation $\tilde{\mathbf{K}}_{i}$, $\tilde{\mathbf{K}}_j$ and $\tilde{\mathbf{K}}_{(i,k)}$ by $\tilde{\mathbf{K}}:=\mathbf{HKH}$, where $\mathbf{H}$ is the normalizing matrix.
\State Conduction eigenvalue decomposition for each normalized kernel by $\tilde{\mathbf{K}}=\mathbf{V \Lambda V}^T$, where $\mathbf{\Lambda}$ is the diagonal matrix containing the non-negative eigenvalues. 
\State Calculate the conditional kernel function by $\tilde{\mathbf{K}}_{(i,k)|k} = \mathbf{R}_k \tilde{\mathbf{K}}_{(i,k)} \mathbf{R}_{k}$ and $\tilde{\mathbf{K}}_{j|k}=\mathbf{R} \tilde{\mathbf{K}}_{j} \mathbf{R}_k$, where $\mathbf{R}_k= \epsilon (\tilde{\mathbf{K}}_k + \epsilon \mathbf{I})^{-1}$ with $\epsilon$ and $\mathbf{I}$ being a small positive regularization parameter and identity matrix, respectively.
\State Calculate p-value based on the statistic $T_{CI} := \frac{1}{n}Tr(\tilde{\mathbf{K}}_{(i,k)|k} \tilde{\mathbf{K}}_{j|k})$, where $Tr$ is the trace of a matrix.

\State \textbf{Output:} The p-value and whether the conditional independence $\mathbf{x}^{i}_{1:t} \indep \mathbf{x}^{j}_{1:t} \mid \mathbf{x}^{k}_{1:t})$ holds.
\end{algorithmic}
\end{algorithm}

\subsubsection{SyPI}
However, directly applying the above-mentioned kernel-based CIT will contribute to plenty of false positives and false negative causal dependencies, which will cause perturbation for the training of the downstream model. To reduce false positive and false negative rates, we apply the SyPI algorithm \cite{mastakouri2021necessary} to further filter the fake relationships generated from kernel-based CIT.  

The algorithm of SyPI is illustrated well in Algorithm 2. Note that the input is a 2D array \textbf{X} (candidate time series) and a vector Y (target), and the output \textbf{$A_{CAU}$} is a set with indices of the time series that were identified as causes, we treat it as our causal adjacency matrix here. min\_lags represent the minimum lag between time series of the candidate node and target node. $\tau$ means the times window.

\begin{algorithm}
\caption{SyPI Algorithm}
\begin{algorithmic}[1] 
\State \textbf{Input:} $\textbf{X}$, $Y$
\State \textbf{Output:} $A_{CAU}$
\State $n_{vars} = shape(\textbf{X}, 1)$, $A_{CAU} = [\ ]$
\State $w = min\_lags(\textbf{X}_{t-w-\tau+1:t-w}, Y)$
\For{$i = 1$ to $n_{vars}$}
    \State $S_i = \bigcup_{j=1, j \neq i}^{n_{vars}} \{X^j_{t-\tau-1:t-1}\}$
    \State $pvalue1 =$

    \Statex \hspace{\algorithmicindent}\hspace{\algorithmicindent} KCIT$(X^i_{t-w-\tau+1:t-w}, Y_{t-\tau+1:t}, [S_i, Y_{t-\tau:t-1}])$
    
    \If{\underline{$pvalue1 < threshold1$}}
        \State $pvalue2 = \text{KCIT}(X^i_{t-w-\tau:t-w-1},$
        \Statex \hspace{\algorithmicindent}\hspace{\algorithmicindent} $Y_{t-\tau+1:t}, [S_i, X^i_{t-w-\tau+1:t-w}, Y_{t-\tau-1:t-1}])$
        \If{\underline{$pvalue2 > threshold2$}}
            \State $A_{CAU} = append(A_{CAU}, X^i_{t-w-\tau+1:t-w})$
        \EndIf
    \EndIf
\EndFor
\end{algorithmic}
\end{algorithm}

\subsubsection{Spatial Pre-selection}
The method above is still $\mathcal{O}(n^2)$ computational complexity. To reduce the computation burden and also incorporate the spatial information. We propose to use the correlation matrix in Eq.~\ref{eq:corr} for the pre-selection of potential causal candidates. For each node $i$, we define the potential causal factor candidates as the neighbouring nodes with the $M$ highest correlations, where $M$ is a hyperparameter to be tuned. Using this pre-selection approach, the computational complexity can be reduced to $\mathcal{O}(Mn)$. 

\subsection{Spatiotemporal Prediction}
For the learned adjacency matrix of a \( A_{CAU} \), we define the normalized Laplacian matrix as \( L = I - D^{-1/2} A_{CAU} D^{-1/2} \), where \( I \) is the identity matrix, and \( D \) is a diagonal degree matrix with each \( D_{ii} \) being the sum of the elements in the \( i^{th} \) row of \( A_{CAU} \). To approximate the graph convolution operator \( *_G \), we employ K-order Chebyshev polynomials, expressed as:

\begin{equation}
    g_\theta *_G x = g_\theta(L) x = \sum_{k=0}^{K-1} \theta_k \left(T_k(\tilde{L})\right) x,
\end{equation}

Here, \( \theta \) in \( \mathbb{R}^K \) represents the vector of polynomial coefficients. The scaled Laplacian \( \tilde{L} \) is calculated as \( \tilde{L} = \frac{2}{\lambda_{\max}} L - I \), with \( \lambda_{\max} \) being the maximum eigenvalue of \( L \). Chebyshev polynomials are recursively defined by \( T_k(x) = 2xT_{k-1}(x) - T_{k-2}(x) \), starting with \( T_0(x) = 1 \) and \( T_1(x) = x \). Using this K-order Chebyshev polynomial approximation allows each node to be updated based on the information from its \( K \) neighboring nodes.

After graph convolution effectively captures neighboring node data in the spatial dimension, we further refine the node signals by adding a standard convolution layer in the temporal dimension. This layer integrates data from neighboring time slices, updating the node signals accordingly. A final 1×D convolution with a nonlinear neural network layer is used to generate the final prediction \( \hat{\mathbf{X}}_{(t-\tau+1):t} \). We employ the mean squared error (MSE) as our loss function:

\begin{equation}
    \mathcal{L} = \frac{|| Y_{(t-\tau+1):t} - \hat{Y}_{(t-\tau+1):t}||^2}{N \tau},
\end{equation}

where \( || \cdot ||^2 \) is the \( \mathbb{L}_2 \) norm.


\section{Experiment}
\label{sec:exp}
In this section, we use a real-world dataset to evaluate the performance of our proposed method. Specifically, we demonstrate our proposed method as follows:

\subsection{Dataset}
To model human mobility patterns during the COVID-19 pandemic, we employ the SafeGraph dataset (Table \ref{table:dataset}), which aggregates location information from mobile applications on individuals' smartphones. This dataset furnishes comprehensive insights into the migration of populations across various districts such as residential, commercial, and recreational sectors, spanning a broad geographical expanse with precise spatial and temporal resolutions.

We aggregate the SafeGraph data into a weekly time frame since it provides a good balance between
granularity and data availability. This temporal aggregation is meticulously aligned with the periodicity of COVID-19 case reporting. In our preprocessing efforts, we extract critical attributes from the SafeGraph dataset, including visitor counts, duration of stays, and the distances traversed among various locations.

Our dataset encapsulates the aggregated data on weekly visits and additional pertinent attributes for 172 areas in New York, recorded from August 10, 2020, to April 18, 2022. We utilize the z-score normalization method to standardize the visitor metrics, preparing the data for downstream modeling inputs.

In addition to the SafeGraph data, we enhance our analysis by incorporating COVID-19 epidemiological data from the Centers for Disease Control and Prevention (CDC), which includes weekly statistics on cases, hospitalizations, and mortality rates across the United States. This integration is crucial for examining the interplay between human mobility and the dynamics of COVID-19 outbreaks. Complementing our dataset, we also include contextual variables such as income levels, population density, and points of interest from open data sources in New York City, thereby enriching our analytical framework.

\begin{table}
\small 
\begin{threeparttable}
\caption{Summary of SafeGraph datasets}
\label{table:dataset}
\setlength\tabcolsep{6pt} 
\begin{tabularx}{\columnwidth}{cc}
\toprule
\multicolumn{2}{c}{\textbf{SafeGraph}} \\
\midrule
\# of ZIP codes & 172 \\
\# of Weeks & 90 \\
Time Span & 2020/08/03 - 2022/04/25 \\
\midrule
\multicolumn{2}{c}{\textbf{Point of Interest (POI)}} \\
\# of POIs & 18,912 \\
Types of POIs & \multicolumn{1}{>{\hsize=\hsize}X}{residential(16\%), education(20\%), etc.} \\
\midrule
\multicolumn{2}{c}{\textbf{Median Household Annual Income}} \\
Range & \$31,536 - \$243,571 \\
\midrule
\multicolumn{2}{c}{\textbf{ZIP Code Population}} \\
Range & 1,783 - 111,344 \\

\bottomrule
\end{tabularx}
\end{threeparttable}
\end{table}

\subsection{Experiment Setting}
In our study, the experimental framework revolves around a basic graph convolutional neural network (GCN) model. The
model is designed to predict the mobility of future weeks of a certain node given the last four weeks' data: images from the dataset and vehicle speed data. The choice of a GCN model was intentional, as it allows for easier training and a clearer explanation of how the quality of the adjacency affects model performance.

\subsubsection{Setting for SyPI}
In our experiment, we set the time series length as 50 weeks and the fixed min\_lag as 1 since our time granularity is on a weekly basis, which is quite coarse. We used $threshold1=0.1$  and $threshold2=0.08$ for SyPI. For the configuration in KCIT, we set kernel size $\sigma$ as 10 and kernel width $\sigma_x$ as 0.8 while other hyperparameters remain the default values. 

\subsubsection{Baseline Method}

In our study, we evaluate the effectiveness of the SyPI adjacency matrix in predicting future mobility and compare it to three well-established relationship measurement methods on our dataset. Our baseline adjacency matrix is:
\begin{itemize}
    \item \textbf{Distance matrix}: We calculated the distance between each node through the Google Map API and set the average distance as a threshold to compute the final adjacency matrix.
    \item \textbf{Correlation matrix}: We employed the Pearson Correlation algorithm to calculate the correlation between each node and computed the adjacency matrix by setting the correlation threshold to 0.75 and the p-value threshold to 0.05.
    \item \textbf{Attention matrix}: The attention matrix is obtained from a well-trained GAT model and extracting its attention parameters. For this matrix, we did not set any threshold but kept the same edges as the SyPI adjacency matrix according to the attention value.
\end{itemize}

\subsubsection{Dataset Splitting}
In preparation for training the GCN model, we split the first 74 weeks of the whole 90 weeks data as our in-distribution train/validation dataset and treated the last 16 weeks of data of our dataset which represents the peak of the COVID-19 outbreak as our out-of-distribution test dataset.

\subsubsection{GCN Structure}
Our Spatiotemporal GCN model uses two spatial convolution layers as the encoder and one temporal convolution layer as the decoder. During the training process, the GCN model receives 4 consecutive weeks' data and returns the following 4 weeks' data as the forecasting results.

\subsection{Results}
\subsubsection{Performance Comparison}


Given different random seeds, as shown in Table \ref{table:result}, our CAL method could achieve 14.23\%, 15.49\%, and 50.27\% Root Mean Square Error (RMSE) improvement in predicting the next week's mobility in New York, compared with the results based on distance, correlation, and attention matrix separately. 
The average RMSE improvement of predicting the mobility after 4 weeks is 24.71\%, 10.98\%, and 27.26\%, compared with the results using distance, correlation, and attention adjacency matrix separately.

Meanwhile, the number of edges in the CAL adjacent matrix is reduced by 9.67\% and 15.98\%, compared with the correlation and attention adjacent matrix separately, which verifies the success of filtering the false positive relationships in the distance and correlation adjacency matrix. Furthermore, it increases the sparsity of the adjacent matrix, optimizing the computation time and space complexity during the training and testing progress.

\begin{table}
\centering
\caption{Performance metrics of GCN prediction on future 1-4 weeks based on different adjacency matrices. Bolded represents the best result and underlined means second best.}
\label{table:result}
\small
\setlength{\tabcolsep}{3pt}
\begin{tabular}{lccccc}
\toprule
\multirow{2}{*}{\textbf{\#Adjacency}} & \multicolumn{5}{c}{\textbf{RMSE/MAE}} \\
\cmidrule(lr){2-6}
                                    & T+1 & T+2 & T+3 & T+4 & Avg.\\
\midrule
\textbf{Distance}                   & \underline{.221/.049} & .284/.081 & .339/.115 & .414/.171 & .322/.104\\
\textbf{Correlation}                & .224/.050 &  \underline{.258/.066} & .259/.067 & \underline{.337/.114} &  \underline{.273/.074}\\
\textbf{Attention}                  & .381/.145 & .276/.076 & \textbf{.233/.054} & .412/.170 & .334/.111\\
\textbf{CAL(Ours)}                 & \textbf{.189/.036} & \textbf{.239/.057} &  \underline{.239/.057} &  \textbf{.292/.085} & \textbf{.243/.059}\\
\bottomrule
\end{tabular}
\end{table}

\subsection{Visualization}
\begin{figure}
    \centering
     \subfloat[\centering ][Predict next 1 week (ZIP = 10309)]{{\includegraphics[width=0.8\columnwidth]{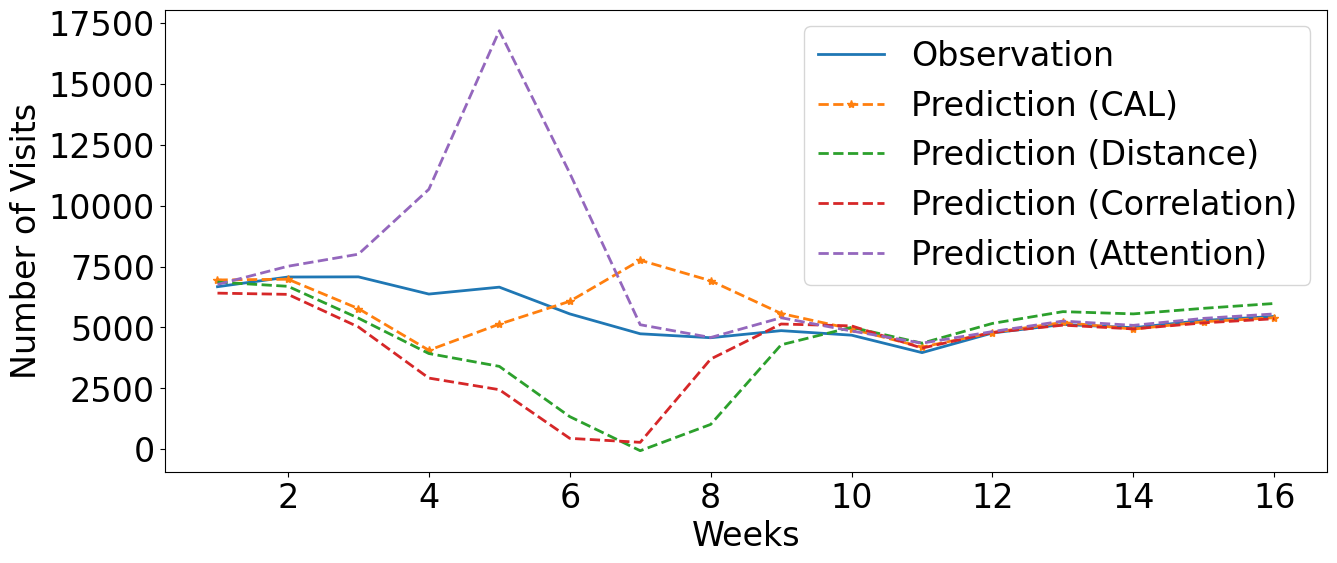} }} 
     \\
     \subfloat[\centering ][Predict next 4 week (ZIP = 10309)]{{\includegraphics[width=0.8\columnwidth]{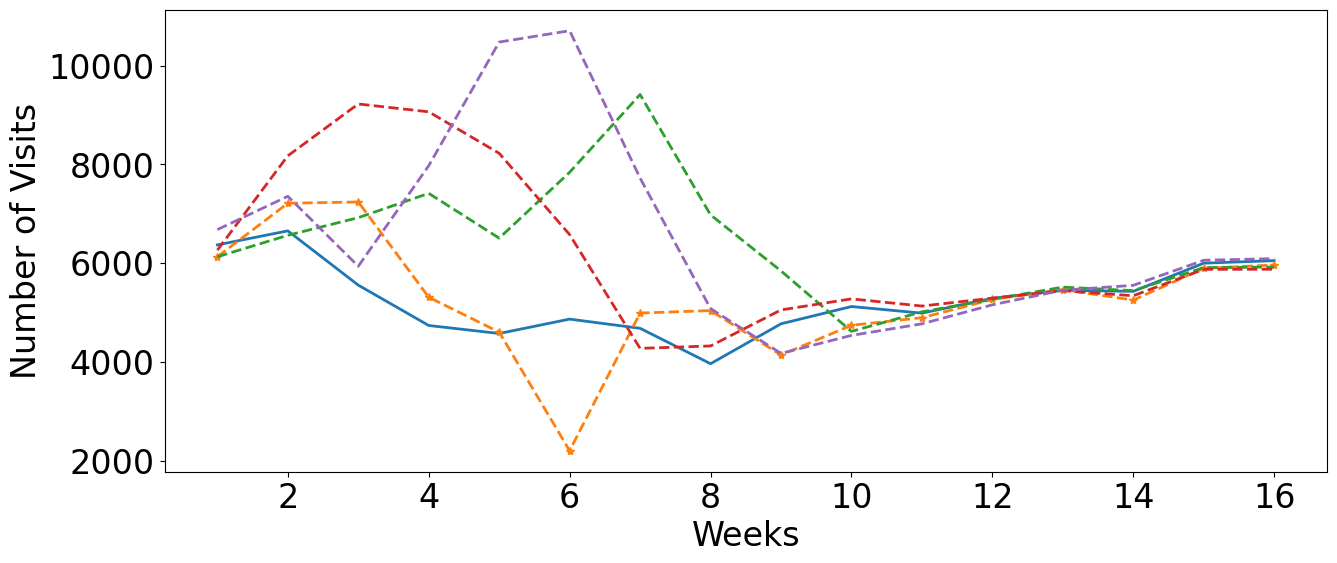} }} 
     \\
     \subfloat[\centering ][Predict next 1 week (ZIP = 10032)]{{\includegraphics[width=0.8\columnwidth]{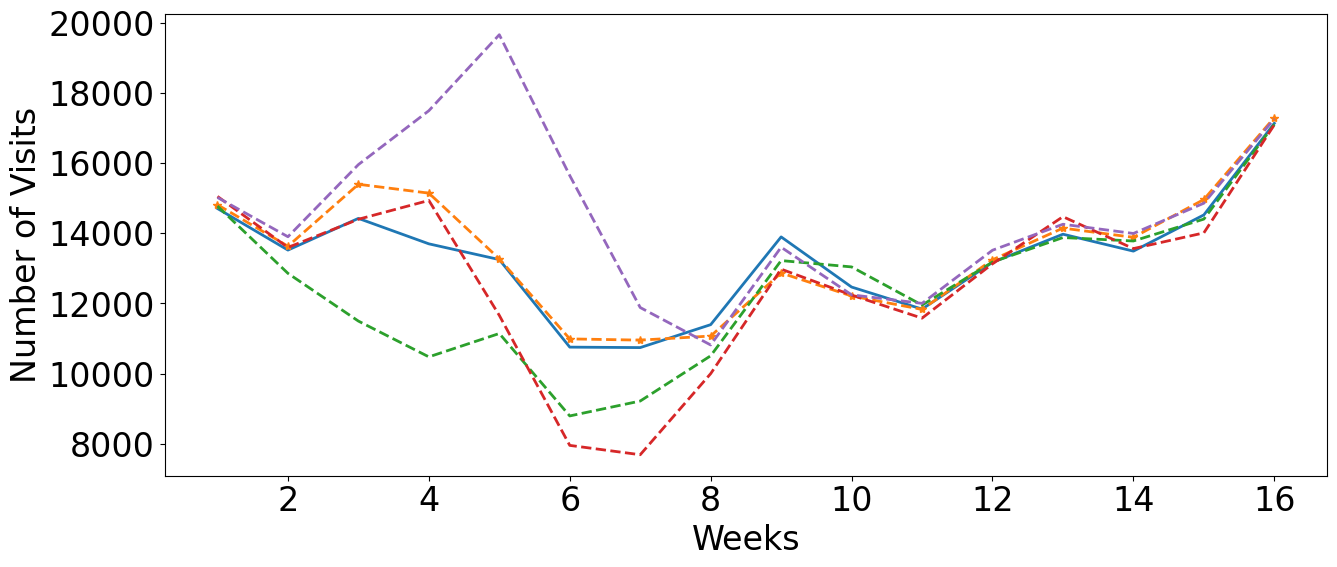} }} 
     \\
     \subfloat[\centering ][Predict next 4 week (ZIP = 10032)]{{\includegraphics[width=0.8\columnwidth]{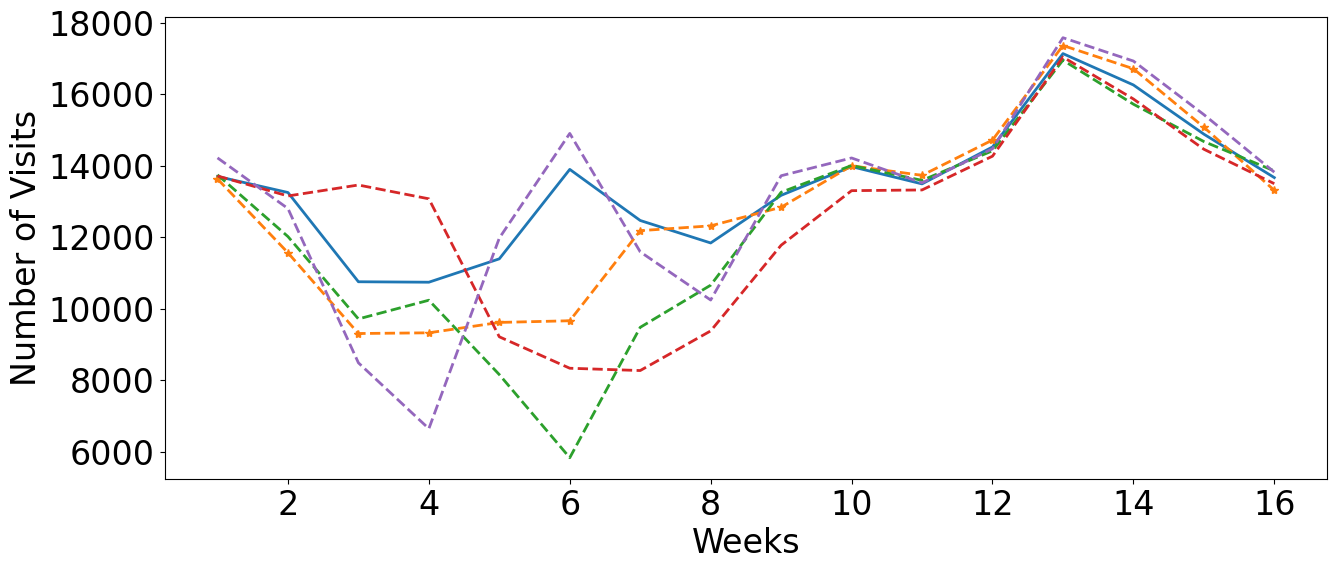} }} 
     \\
    \caption{ GCN prediction of future mobility based on different adjacency matrices for two ZIP codes. }
    \label{fig:prediction}
\end{figure}

The prediction of the CAL is illustrated in Fig \ref{fig:prediction}. The CAL model demonstrates superior performance in fitting the ground truth compared to other models, for both one-week and four-week predictions. Note that there was a COVID-19 outbreak around the sixth week (February 2022) in our test dataset, which provides a strong OOD property and difficulty for data forecasting. However, our method delivers more accurate predictions in that week, even for long-time forecasting, which implies that the CAL model can catch more potential and sustainable causal relationships than other methods.

\begin{figure}%
    \centering
     \subfloat[\centering ]{{\includegraphics[width=1\columnwidth]{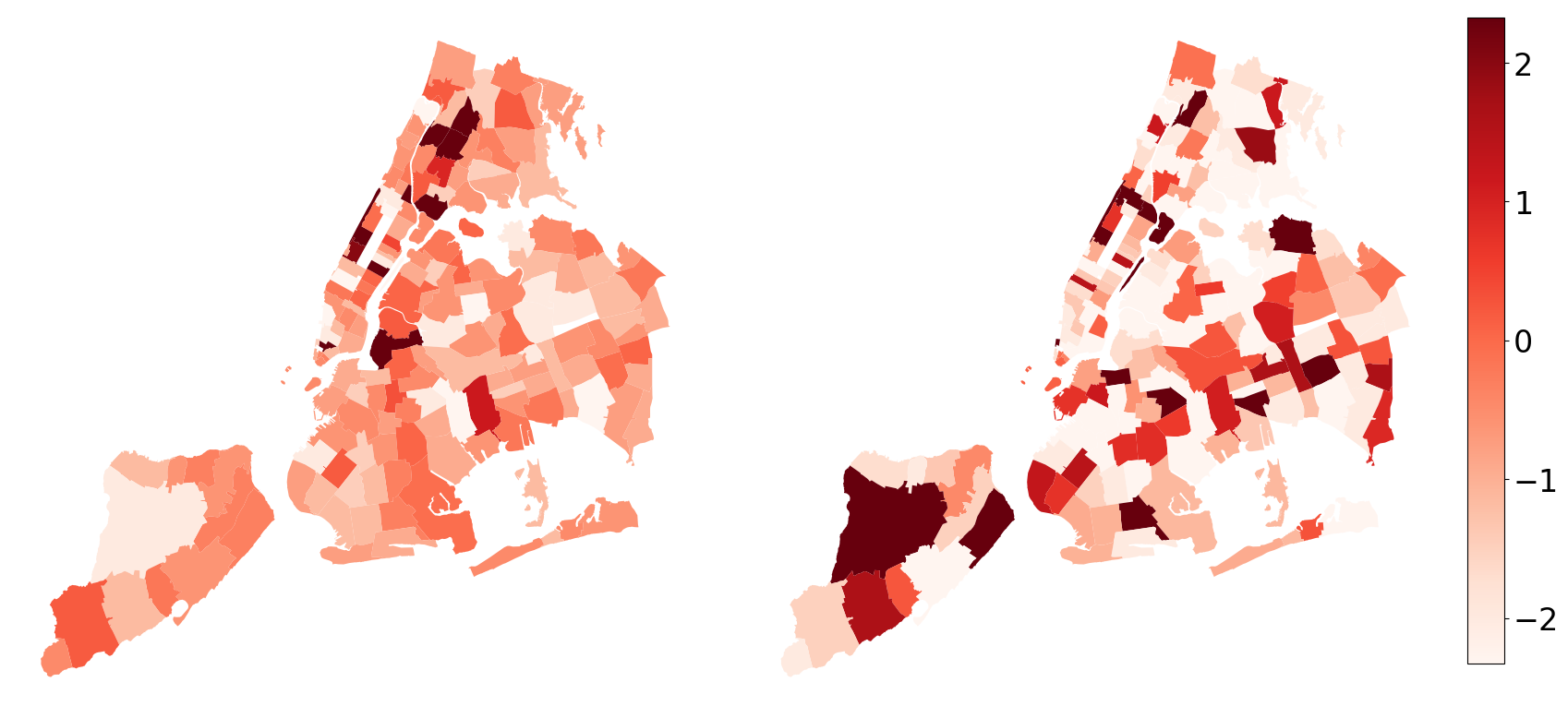} }}
    \caption{ Row (left) and column (right) aggregation of $A_{CAU}$.}%
    \label{fig:visualization}%
\end{figure}

We aggregate the adjacency matrix in the row and column directions and plot these aggregations on the map, as shown in Fig \ref{fig:visualization}, The row and column aggregations represent the total impact received from and sent to other regions, respectively. The figure clearly illustrates that Manhattan exerts the most significant influence on other regions, while Staten Island has the minimal impact, reflecting their respective levels of prominence. Conversely, Staten Island is the most affected by external influences from other regions, which corresponds with expectations.

\section{Conclusions}
\label{sec:conclu}
In this paper, we propose a causal adjacency learning framework that identifies causal factors for each node in a graph. By encoding the results into the adjacency matrix, this causal information can be utilized in other downstream tasks. We demonstrate with real-world datasets that our learned causal adjacency matrix enables models to capture out-of-distribution (OOD) patterns, even if causal learning is not explicitly performed in the downstream task. Additionally, by visualizing the causal adjacency matrix on a geospatial map, we provide a practical interpretation of our method's performance.

This method could be enhanced in two key ways. First, we are considering incorporating information from spatial dimensions beyond using a correlation matrix for pre-selection. Second, we plan to explore the application of our proposed method in various cities that exhibit different patterns.

\bibliographystyle{IEEEtran}
\bibliography{ieee.bib}

\end{document}